Simulating time to event prediction with spatiotemporal echocardiography deep learning

Rohan Shad[1], Nicolas Quach[1], Robyn Fong[1], Patpilai Kasinpila[1], Cayley Bowles[1], Kate M. Callon[1], Michelle C Li[1], Jeffrey Teuteberg[2,3], John P. Cunningham[4], Curtis P. Langlotz[3,5], **William Hiesinger[1,3]**

[1]Department of Cardiothoracic Surgery, Stanford University; [2]Department of Cardiovascular Medicine, [3]Stanford Artificial Intelligence in Medicine Center, [4]Department of Statistics, Columbia University, [5]Department of Radiology and Biomedical Informatics, Stanford University

## Abstract

Integrating methods for time-to-event prediction with diagnostic imaging modalities is of considerable interest, as accurate estimates of survival requires accounting for censoring of individuals within the observation period. New methods for time-to-event prediction have been developed by extending the cox-proportional hazards model with neural networks. In this paper, to explore the feasibility of these methods when applied to deep learning with echocardiography videos, we utilize the Stanford EchoNet-Dynamic dataset with over 10,000 echocardiograms, and generate simulated survival datasets based on the expert annotated ejection fraction readings. By training on just the simulated survival outcomes, we show that spatiotemporal convolutional neural networks yield accurate survival estimates.

## Introduction

Recent advances in diagnostic echocardiography AI systems have shown that not only is it possible to automate standard measures of cardiac function (ejection fraction), but even measure common biomarkers that are typically never measured via echocardiography scans (Hemoglobin, blood urea nitrogen).[1–3] Our group has shown that predicting post-operative right ventricular failure in the setting of left ventricular assist device implantation is feasible by analyzing the full spatiotemporal density of information within each pre-operative echocardiogram.[4] Others have now shown that predicting 1-year mortality is possible using just baseline echocardiography.[5] Though this was developed using one of the largest echocardiographic datasets, the work was limited by the use of potentially under-parameterized neural networks, and treatment of right censored survival data as a binary event.

Time-to-event predictions rather than simple binary predictions will find more relevance in the clinical setting when working with 'prognostic' rather than 'diagnostic' AI systems.[6] Time-to-event analyses are characterized by the ability to predict event probabilities as a *function* of time, whereas binary classifiers can only provide predictions for one predetermined duration. One of the advantages of this is that it accounts for censoring of individuals within the observation period.[7] This more accurately represents the clinical ground truth where individuals may be either lost to follow up, or have yet to develop an event of interest.

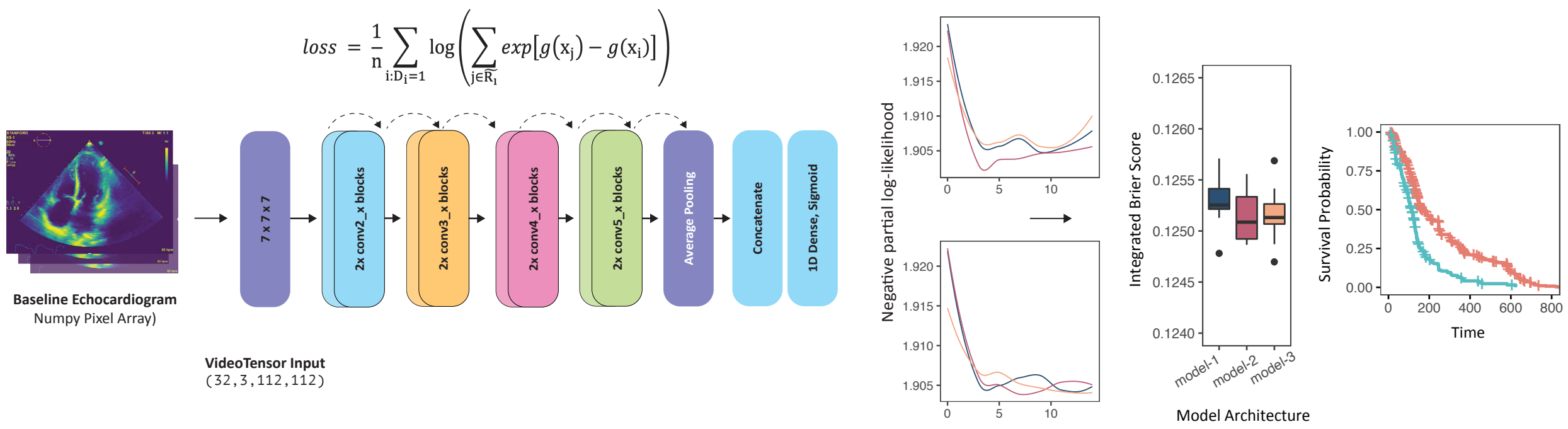

**Figure 1:** Echocardiography video input is fed into 3-dimensional 18-layer residual networks with varying degrees of spatiotemporal coupling. The model is trained using an implementation of the cox proportional loss function (DeepSurv) to output survival estimates

The cox-proportional hazards model is a commonly utilized form of time-to-event analysis, and numerous attempts to extend the standard cox model using deep learning have been described.[8,9] Kvamme *et. al.* proposed a series of proportional and non-proportional extensions of the Cox model, with additional implementations of survival methods described in the past such as *DeepSurv* and *DeepHit*.[9] Integrating cox-models with computer vision systems has shown promise in predicting cancer outcomes from histopathological slides.[8] Given the ubiquitous nature of echocardiography in the screening, management, and diagnosis of heart disease, the argument for extending capabilities of echocardiography AI systems with time-to-event analyses is a logical one.[10–15] Unfortunately. large public datasets of baseline echocardiographs with paired long-term outcomes do not yet exist.

To explore the feasibility of these methods when applied to deep learning based echocardiography analyses, in this paper we utilize the Stanford EchoNet-Dynamic dataset with over 10,000 echocardiograms, and generate simulated survival datasets based on the expert annotated ejection fraction readings.[1] We hypothesize that without explicitly training neural networks to regress ejection fraction, our ML system will yield accurate predictions of survival outcomes. Figure 1 shows an overview of our project.

## Methods:

### *AIMI Echo dataset*

We use the recently released Stanford EchoNet dataset – a dataset with 10030 echocardiography videos and paired expert annotations and measurements of the left ventricle at end systole and diastole.[1] We use an input resolution of 112 x 112 pixels and 32 frames per video. We skip every alternate frame in each video while pre-processing, allowing for greater temporal coverage within each subsample of 32 frames. The dataset consisted of 7465, 1288, and 1277 videos in the training, validation, and testing datasets respectively.

### *Simulation of survival data:*

We used the coxed R package to simulate ten survival datasets a user defined covariate. Briefly, we generate baseline hazard functions using the flexible hazards method.[16] In an effort to generate realistic survival data, the coefficient $\beta$ was set to 0.035, to approximate observed long-term survival in participants of the Framingham Heart Study.[10] Simulated survival for each observation is calculated based on the baseline hazard function, ejection fraction (Taken from the AIMI EchoNet dataset) , and the user supplied coefficient. We specify 15% of the dataset to be right-censored independent of any input covariates, with the hazard functions specified as a step function. The baseline hazard functions are iteratively generated for each simulated dataset within these constraints. In Figure 2 we show the simulated baseline functions across four simulated datasets. Each simulated dataset has a unique baseline hazard and survivor function.

### *Neural Network architectures:*

We evaluated a series of 3-dimensional convolutional neural networks with varying degrees coupling between the spatial and temporal streams. The three Kinetics400 pre-trained residual network architectures are adapted from the pytorch torchvision models repository, and are constructed from basic (non-bottlenecked) residual blocks.[17,18] The tested networks are each 18-layers deep: a 3D residual network (r3d_18), mixed convolutional network where initial layers are 3D convolutional layers followed by higher level 2D layers (mc3_18), and a decomposed 2D convolutional spatial network with alternate 1D temporal convolutions.[19] Each network was configured to terminate in a single output node that estimates the risk function parameterized by network weights. We used the Stochastic Gradient Descent optimizer with momentum set to 0.9 and weight decay set to 1e-4 for all the experiments. Learning rate was set to 5e-4. The models were built using the Pytorch deep learning library and distributed experiments configured using the Pytorch Lightning framework.[20] Each echocardiography video file was processed as a tensor of size (frames, channels, height, width) $[32,\ 3, 112, 112]$. Random rotations of $\pm$ 5 degrees were used for train-time video augmentation.

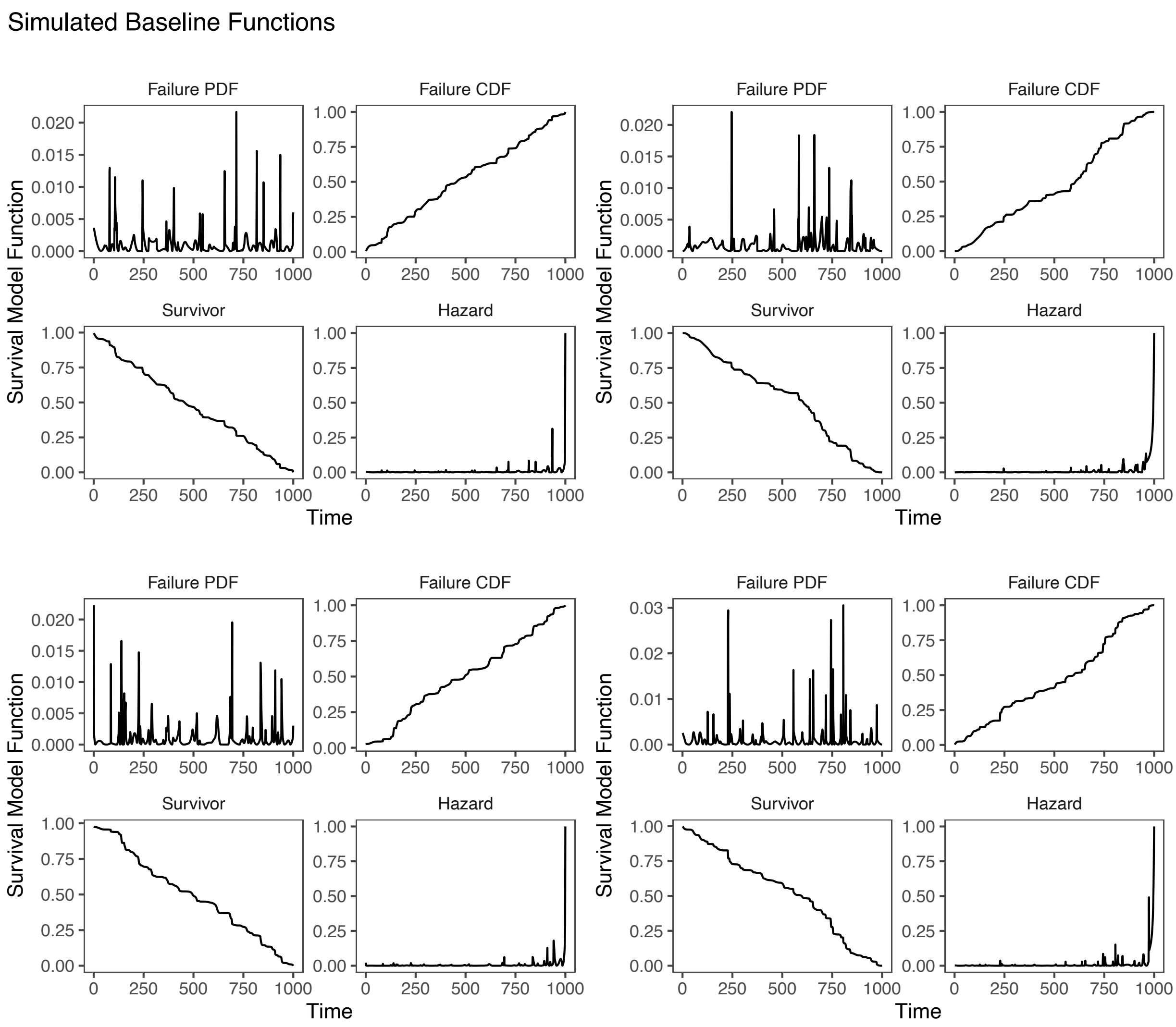

**Figure 2:** Selection of simulated datasets each with unique survivor and hazard functions, based on the constraints provided. We generate 10 such simulated datasets for the entire 10030 patient dataset in the AIMI EchoNet dataset

*Survival model framework*

For all tested network architectures, we use an implementation of the *DeepSurv* Cox-proportional hazards loss function – a negative partial log likelihood derived from the cox partial likelihood with Breslow's method for handling tied event times.[8] Since the entire training dataset cannot fit on GPU memory, we utilize the batched approximation of the negative partial log-likelihood loss function optimized via gradient descent.[9] The implementation of *DeepSurv* used in this paper is a refactored version of the loss function provided in the *pycox* python package and described in detail by Kvamme *et al*.[9] The loss function is given by:

$$loss = \frac{1}{n}\sum_{i:D_i=1} log\left(\sum_{j\in \widetilde{R_i}} exp[g(x_j) - g(x_i)]\right)$$

Predictions are obtained by estimating the survival functions for the test dataset. This is calculated by first calculating the baseline hazards $H_0(t)$, and then the relative risk function $exp[g(x)]$. Other non-parametric cox models, and discrete time models were not evaluated as part of this work.

$$H(t|x) = \int_0^t h_0(s)exp[g(x)]ds = H_0(t)exp[g(x)]$$

*Experimental design:*

A total of 30 permutations of simulated dataset number and model architecture were configured using a cluster optimized script in pytorch lightning. Each model was trained 3 times on each simulated training datasets, and concordance index and brier scores were calculated at the end of each epoch on the validation dataset. Models were trained for 15 epochs, model weights were frozen that yielded the lowest validation loss. Concordance index and integrated brier scores were used to assess the performance of each model on the test dataset. The calculation of these metrics have been described previously.[9,21] Briefly, the concordance index estimates the probability of predicted survival times for two random individuals being ordered the same as their true survival.[21] The brier score on the other hand, represents the average squared distance between observed and predicted survival probability at a given time *t.* The squared distances are weighted with an inverse probability of censoring weights method.[22] Integrating the brier scores over all available times gives us the Integrated Brier Score – a number between 0 to 1, with 0 being the best possible value. The experiments were conducted on the Stanford Sherlock High Performance Computing Cluster, on a total of 32 Nvidia 2080Ti GPUs across all experiments.

Results

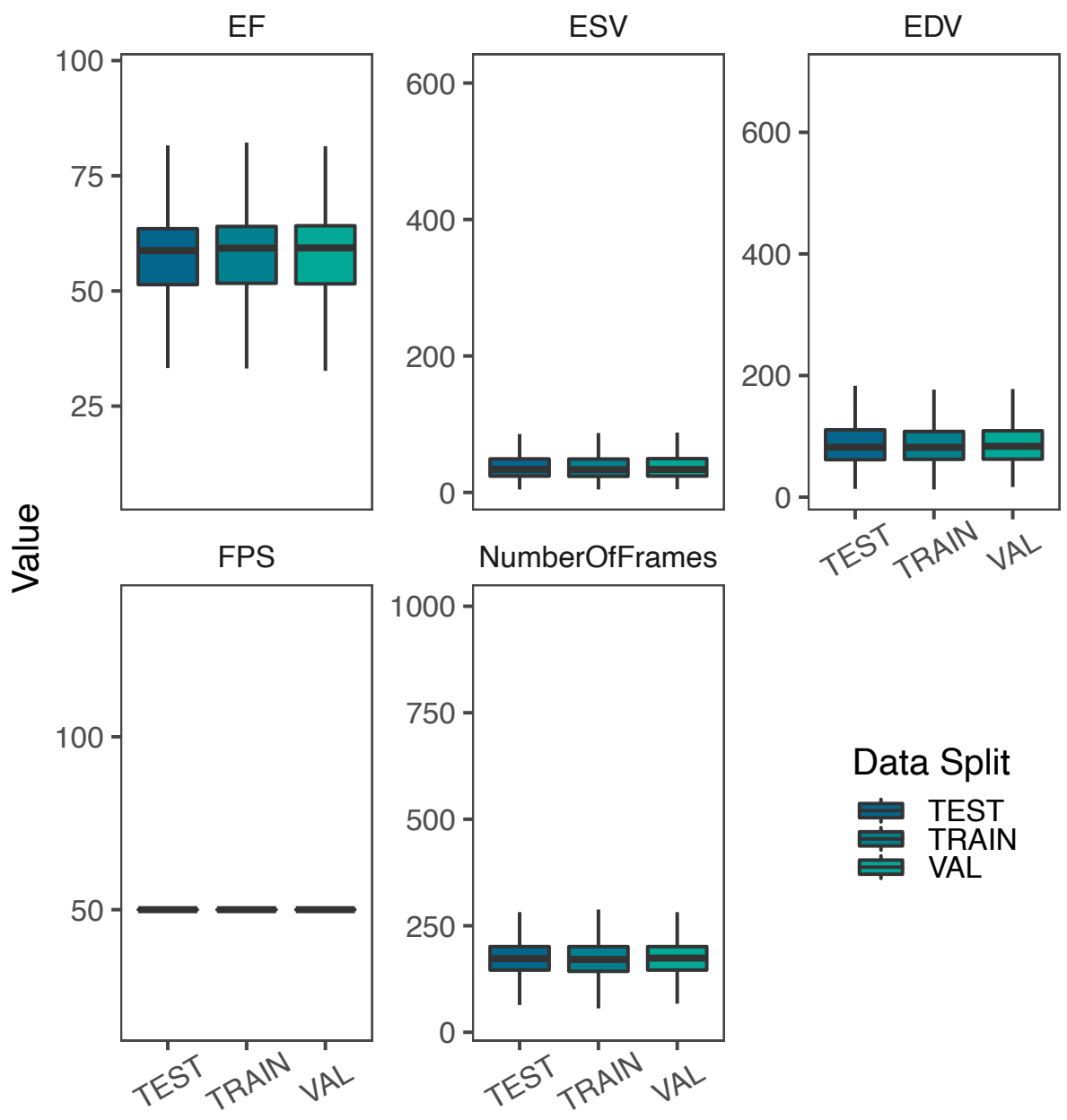

**Figure 3:** Summary of measurements provided as part of the EchoNet dataset by training, validation, and testing data splits

The EchoNet dataset has been described previously.[1] In Figure 3 we show the distribution of key variables across the dataset. The measurements are show similar distributions across all three datasets. The mean ejection fraction was 55.77 ± 12.40%, 55.82 ± 12.30%, 55.49 ± 12.23% in the training, validation, and testing dataset respectively. The training curves are shown in Figure 4a, for each combination of simulated dataset and architecture choice, the training loss (negative partial log likelihood) converges rapidly within the first few epochs. We let the models train for 15 epochs each, with model checkpoints saved for the best validation loss. After freezing the models, we evaluate each model on the testing dataset for each iteration. Figure 4b details the distribution of concordance indices and integrated brier scores across each network architecture. There were minimal differences in performance across the three models, though we did not explore additional hyperparameters. We additionally calculated the concordance index for each dataset by fitting a standard cox-proportional hazards model with ejection fraction as the predictor using the survival package in R (v3.6.3).

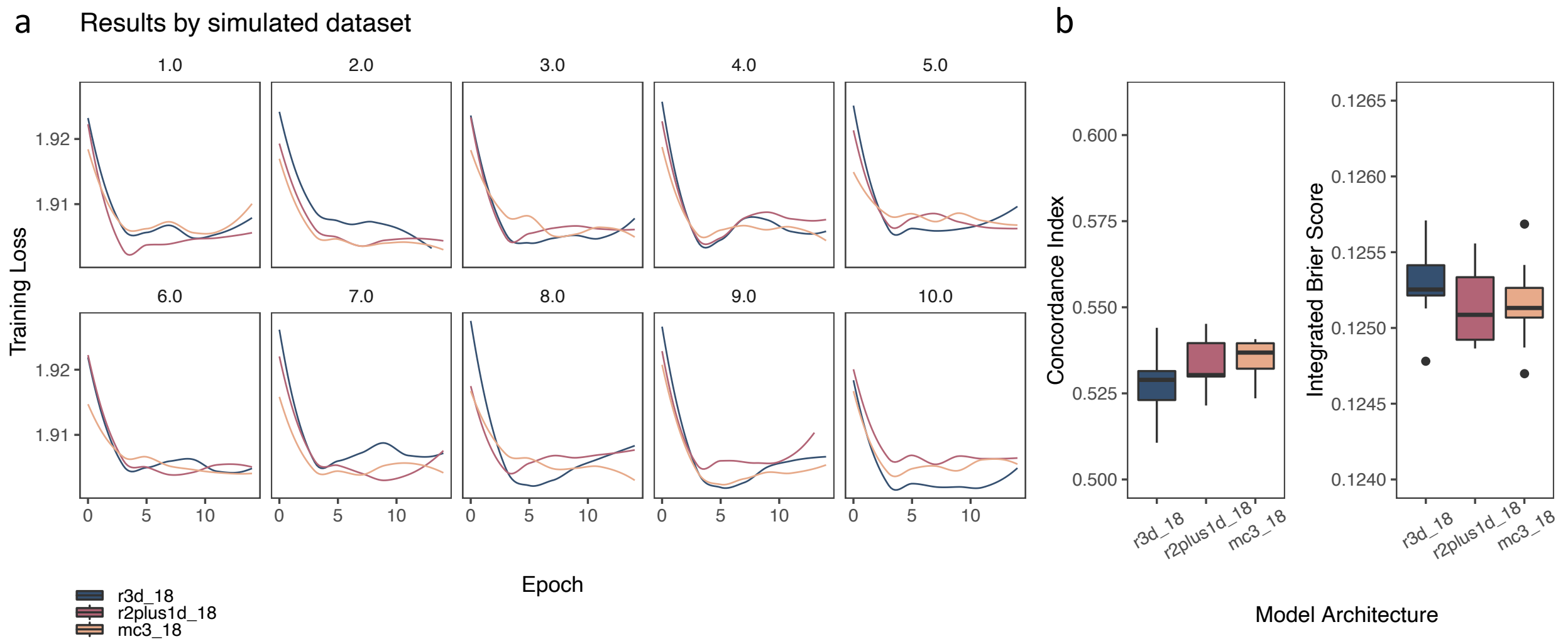

**Figure 4a**: Training curves for each model architecture: r3d_18, r2plus1d_18, mc3_18 across ten simulated survival datasets. **b.** Box-plots illustrating the distribution of Concordance Index and Integrated Brier scores across all the experiments.

To define the ability of the spatiotemporal ResNets to estimate the true predictor via training on survival outcomes, we plot the predicted individual part-log likelihoods to the true individual partial-log likelihoods. We display a selection of these plots across four simulated datasets in Figure 5, showing that the estimates are very close to the true likelihoods.

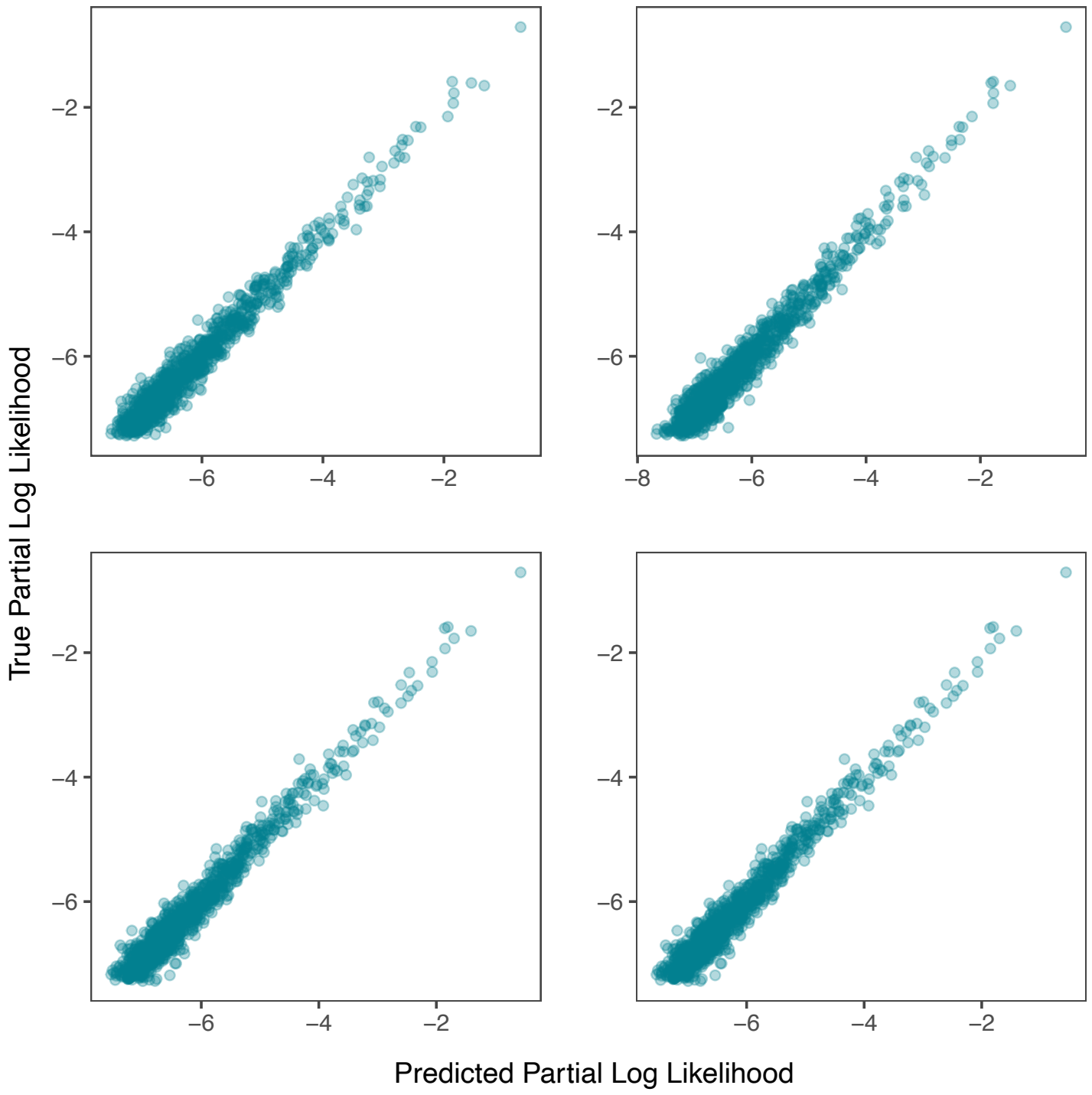

**Figure 5:** Plots for predicted and true partial log-likelihoods across four simulated datasets.

## Discussion

The prevalence of Asymptomatic left ventricular dysfunction (ALVD) is estimated to be between 3 – 8% with a higher prevalence in men.[10] ALVD is also twice as common as symptomatic heart failure.[23] Reports from the Framingham Heart Study have shown that individuals with ALVD are at high risk of heart failure and death, even with mild impairment of contractile function on baseline echocardiography.[10] Visual estimates of systolic function have historically been used to distinguish normal from abnormal LV systolic function, however characterizing diastolic dysfunction requires a more comprehensive echocardiographic assessment.[10,24] Finally, the STOP-HF randomized control trial demonstrated a significant reduction in newly diagnosed heart failure and emergency cardiovascular hospitalizations, by screening and treating individuals at risk of heart failure.[25]

Integrating survival analyses into echocardiography AI systems may help with efforts to identify high risk individuals and direct them towards more intensive follow-up and preventing pharmacological therapy. Such efforts of course, would require large population-based echocardiography data. The Framingham Heart Study (NHLBI) – the largest prospectively followed cohort of participants, now in its 3rd generation, was recruited for identifying common characteristics that contribute to cardiovascular disease. Each of the 15537 participants were comprehensively evaluated at baseline during recruitment into the study with a physical exam and echocardiogram. Furthermore, each participant was evaluated at regular intervals over decades allowing one to accurately assess the progression of 'asymptomatic' disease to outcomes of interest such as overt congestive heart failure or death. While the earliest baseline echocardiographs featured only M-mode echocardiography scans, acquisition of many standard echocardiographic views (A4C / PLAX / PSAX views.) began soon after. There are efforts underway to digitize these scans to enable deep learning research.[26] Such resources are critical to the development of translatable AI systems.

When applied to large real-world prospective datasets, it is possible that additional features may be detectable in echocardiography videos predictive of death, such as ventricular or atrial motion features.[5] As shown in Figure 5, the methods outlined in this manuscript make learning such features without explicit and direct supervision feasible. The ability to predict survival curves for each patient is powerful, allowing for risk stratification based on myocardial motion features directly instead of relying on human-derived metrics of cardiac function. Aside from population-based screening for clinically significant ALVD, deep learning-based risk stratification may be used to inform the management of patients with varying grades of heart failure, pulmonary hypertension, and valve disease. Combined with discriminative saliency maps, it may further be possible to highlight structural areas of significance for these deep learning-based predictions.[27–29]

The strengths of this study are the use of a constrained simulation approach to generate physiologically plausible survival datasets based on real world echocardiography files. We evaluated our models across 10 generated baseline hazard functions, showing that our results are robust to differences in baseline hazards. All three residual network architectures performed similarly with the same hyperparameters with respect to concordance and integrated brier scores. While the original echocardiography files are sourced from real world patients, key demographic variables such as age, gender, race and ethnicity were unfortunately never publicly released.[1] As a result, our models have no implicit ability to adjust baseline survival by patient age or gender, and neither are we able to define if our models are systematically biased towards any demographic. Including structured tabular input variables within fusion models (combining both video and tabular input) may help adjust for some of these variables, in addition to providing opportunities to better explain the relative importance of demographic variables and baseline echocardiography videos.[5,30]

The simulation parameters we used restricted the generation of survival functions that resembled real world clinical and population data. As such the results here should not be taken as a general validation of the underlying methods. In certain applications, the proportional hazards assumption may not hold true. Survival functions without a proportional hazards constraint may provide superior results in such scenarios.[9,31] Finally, we did not extensively study the impact of model architecture on performance beyond the three 18-layer residual networks described in this paper. In summary, we show that accurate predictions of survival estimates can be made by using deep learning extensions of the cox-proportional hazards loss with video based convolutional neural networks. These methods when applied to time resolved cardiovascular imaging, may find significant utility in the screening and management of individuals with cardiovascular disease.

Acknowledgements:

Some of the computing for this project was performed on the Stanford Sherlock. We would like to thank Stanford University and the Stanford Research Computing Center for providing computational resources and support. This project was in part supported by a Stanford Artificial Intelligence in Medicine (AIMI) Seed Grant. We would like to acknowledge Håvard Kvamme for assisting with refactoring the survival loss functions to pytorch.